\documentclass[sigconf]{acmart}
\AtBeginDocument{%
  \providecommand\BibTeX{{%
    \normalfont B\kern-0.5em{\scshape i\kern-0.25em b}\kern-0.8em\TeX}}}

\copyrightyear{2021}
\acmYear{2021}
\setcopyright{iw3c2w3}
\acmConference[WWW '21]{Proceedings of the Web Conference 2021}{April 19--23, 2021}{Ljubljana, Slovenia}
\acmBooktitle{Proceedings of the Web Conference 2021 (WWW '21), April 19--23, 2021, Ljubljana, Slovenia}
\acmPrice{}
\acmDOI{10.1145/3442381.3449822}
\acmISBN{978-1-4503-8312-7/21/04}

\usepackage{graphicx,wrapfig}
\usepackage{subfigure}
\usepackage{enumitem}
\usepackage{pifont}
\usepackage{graphicx}
\usepackage{float}
\usepackage{multirow}
\usepackage[ruled]{algorithm2e}
\usepackage{color}
\definecolor{grey}{RGB}{211,211,211}
\newcommand{\projtitle}{SUGAR}
\newcommand{\projtitleRL}{SUGAR-FixedK}
\newcommand{\projtitleMI}{SUGAR-NoMI}
\newcommand{\projtitleMICor}{SUGAR-MICorrupt}
\begin{document}

\title{SUGAR: Subgraph Neural Network with Reinforcement Pooling and Self-Supervised Mutual Information Mechanism}

\author{Qingyun Sun{$^{1,2}$}, Jianxin Li{$^{1,2}$}, Hao Peng{$^{1}$}, Jia Wu{$^{3}$}, Yuanxing Ning{$^{1,2}$}, Phillip S. Yu{$^{4}$}, Lifang He{$^{5}$}}
\affiliation{
  \institution{
   $^1$
   Beijing Advanced Innovation Center for Big Data and Brain Computing, Beihang University, Beijing 100191, China\\
   $^2$
   School of Computer Science and Engineering, Beihang University, Beijing 100191, China\\
   $^3$
   Department of Computing, Macquarie University, Sydney, Australia\\
   $^4$
   Department of Computer Science, University of Illinois at Chicago, Chicago, USA\\
   $^5$
   Department of Computer Science and Engineering, Lehigh University, Bethlehem, PA, USA
   }
  \country{}
}

\email{{sunqy,lijx,penghao,ningyx}@act.buaa.edu.cn, jia.wu@mq.edu.au, psyu@uic.edu, lih319@lehigh.edu}
\renewcommand{\authors}{Qingyun Sun, Jianxin Li, Hao Peng, Jia Wu, Yuanxing Ning, Phillip S. Yu, Lifang He}
\renewcommand{\shortauthors}{Sun et al.}

\begin{abstract}
Graph representation learning has attracted increasing research attention. 
However, most existing studies fuse all structural features and node attributes to provide an overarching view of graphs, neglecting finer substructures' semantics, and suffering from interpretation enigmas. 
This paper presents a novel hierarchical subgraph-level selection and embedding-based graph neural network for graph classification, namely SUGAR, to learn more discriminative subgraph representations and respond in an explanatory way. 
SUGAR reconstructs a sketched graph by extracting striking subgraphs as the representative part of the original graph to reveal subgraph-level patterns. 
To adaptively select striking subgraphs without prior knowledge, we develop a reinforcement pooling mechanism, which improves the generalization ability of the model. 
To differentiate subgraph representations among graphs, we present a self-supervised mutual information mechanism to encourage subgraph embedding to be mindful of the global graph structural properties by maximizing their mutual information. 
Extensive experiments on six typical bioinformatics datasets demonstrate a significant and consistent improvement in model quality with competitive performance and interpretability. 
\end{abstract}



\maketitle

\section{Introduction}
Graphs have been widely used to model the complex relationships between objects in many areas including computer vision~\cite{felzenszwalb2004efficient,yang2018graph}, natural language processing~\cite{peng2018large}, anomaly detection~\cite{noble2003graph,akoglu2015graph}, academic network analysis~\cite{zhang2019oag,sun2020pairwise}, bioinformatics analysis~\cite{Toivonen2003Statistical,Dobson2003Distinguishing}, etc. 
By learning a graph-based representation, it is possible to capture the sequential, topological, geometric, and other relational characteristics of structured data. 
However, graph representation learning is still a non-trivial task because of its complexity and flexibility. 
Moreover, existing methods mostly focus on node-level embedding~\cite{kipf2016semi,velivckovic2017graph,hamilton2017inductive}, which is insufficient for subgraph analysis. 
Graph-level embedding~\cite{shervashidze2011weisfeiler,Niepert2016Learning,zhang2018end,ying2018hierarchical} is critical in a variety of real-world applications such as predicting the properties of molecules in drug discovery~\cite{ma2018drug}, and community analysis in social networks~\cite{li2019semi}. 
In this paper, we focus on graph-level representation for both subgraph discovery and graph classification tasks in an integrative way.

In the literature, a substantial amount of research has been devoted to developing graph representation techniques, ranging from traditional graph kernel methods to recent graph neural network methods. 
In the past decade, many graph kernel methods~\cite{kriege2020survey} have been proposed that directly exploit graph substructures decomposed from it using kernel functions, rather than vectorization. 
Due to its specialization, these methods have shown competitive performance in particular application domains. 
However, there are limitations in two aspects. 
Firstly, kernel functions are mostly handcrafted and heuristic~\cite{borgwardt2005shortest,shervashidze2011weisfeiler,Shervashidze2009Efficient,Yanardag2015Deep}, which are inflexible and may suffer from poor generalization performance. 
Secondly, the embedding dimensionality usually grows exponentially with the growth in the substructure’s size, leading to sparse or non-smooth representations~\cite{cai2018comprehensive}.

With the recent advances in deep learning, Graph Neural Networks (GNNs)~\cite{wu2020comprehensive} have achieved significant success in mining graph data. 
GNNs attempt to extend the convolution operation from regular domains to arbitrary topologies and unordered structures, including spatial-based~\cite{Niepert2016Learning,Simonovsky2017Dynamic,gilmer2017neural} and spectral-based methods~\cite{bruna2013spectral,defferrard2016convolutional,li2018adaptive}. 
Most of the current GNNs are inherently flat, as they only propagate node information across edges and obtain graph representations by globally summarizing node representations. 
These summarisation approaches include averaging over all nodes~\cite{duvenaud2015convolutional}, adding a virtual node~\cite{Li2015Gated}, using connected layers~\cite{gilmer2017neural} or convolutional layers~\cite{zhang2018end}, etc.

However, graphs have a wide spectrum of structural properties, ranging from nodes, edges, motifs to subgraphs. 
The local substructures (i.e., motifs and subgraphs) in a graph always contain vital characteristics and prominent patterns~\cite{ullmann1976algorithm}, which cannot be captured during the global summarizing process. 
For example, the functional units in organic molecule graphs are certain substructures consisting of atoms and the bonds between them~\cite{debnath1991structure,Dobson2003Distinguishing}. 
To overcome the problem of flat representation, some hierarchical methods~\cite{ying2018hierarchical,gao2019graph} use local structures implicitly by coarsening nodes progressively, which often leads to the unreasonableness of the classification result. 
To exploit substructures with more semantics, some researchers~\cite{monti2018motifnet,yang2018node,lee2019graph,peng2020motif} exploit motifs (i.e., small, simple structures) to serve as local structure features explicitly. 
Despite its success, it requires domain expertise to carefully design specific motif extracting rules for various applications carefully. 
Moreover, subgraphs are exploited to preserve higher-order structural information by motif combination~\cite{yang2018node}, subgraph isomorphism counting~\cite{bouritsas2020improving}, rule-based extraction~\cite{xuan2019subgraph}, etc. 
However, effectively exploiting higher-order structures for graph representation is a non-trivial problem due to the following major challenges: 
(1) \textit{Discrimination}. 
Generally, fusing all features and relations to obtain an overarching graph representation always brings the potential concerns of over-smoothing, resulting in the features of graphs being indistinguishable. 
(2) \textit{Prior knowledge}. 
Preserving structural features in the form of similarity metrics or motif is always based on heuristics and requires substantial prior knowledge, which is tedious and ad-hoc. 
(3) \textit{Interpretability}. 
Many methods exploit substructures by coarsening them progressively. 
This is not suitable to give prominence to individual substructures, resulting in a lack of sufficient interpretability. 

To address the aforementioned challenges, we propose a novel \textbf{SU}b\textbf{G}r\textbf{A}ph neural network with \textbf{R}einforcement pooling and self-supervised mutual information mechanism, named \textbf{\projtitle}. 
Our goal is to develop an effective framework to adaptively select and learn discriminative representations of striking subgraphs that generalize well without prior knowledge and respond in an explanatory way. 
\projtitle~reconstructs a sketched graph to reveal subgraph-level patterns, preserving structural information in a three-level hierarchy: node, intra-subgraph, and inter-subgraph. 
To obtain more representative information without prior knowledge, we design a reinforcement pooling mechanism to select more striking subgraphs adaptively by a reinforcement learning algorithm. 
Moreover, to discriminate subgraph embeddings among graphs, a self-supervised mutual information mechanism is also introduced to encourage subgraph representations preserving global properties by maximizing mutual information between local and global graph representations. 
Extensive experiments on six typical bioinformatics datasets demonstrate significant and consistent improvements in model quality. 
We highlight the advantages of \projtitle\footnote{Code is available at https://github.com/RingBDStack/SUGAR. }
~as follows: 
\begin{itemize}[leftmargin=*]
    \item \textbf{Discriminative.} 
    \projtitle~learns discriminative subgraph representations among graphs, which are aware of both local and global properties. 
    \item \textbf{Adaptable.} 
    \projtitle~adaptively finds the most striking subgraphs given any graph without prior knowledge, which allows it to perform in a superior way across various types of graphs. 
    \item \textbf{Interpretable.} 
    \projtitle~explicitly indicates which subgraphs are dominating the learned result, which provides insightful interpretation into downstream applications. 
\end{itemize}
\section{Related Work}
This section briefly reviews graph neural networks, graph pooling, and self-supervised learning on graphs. 
\subsection{Graph Neural Networks}
Generally, graph neural networks follow a message-passing scheme recursively to embed graphs into a continuous and low-dimensional space. 
Prevailing methods capture graph properties in different granularities, including node~\cite{Niepert2016Learning,taheri2018learning,ivanov2018anonymous}, motif~\cite{monti2018motifnet,lee2019graph,peng2020motif}, and subgraph~\cite{yang2018node,xuan2019subgraph,alsentzer2020subgraph}. 

Several works~\cite{Niepert2016Learning,taheri2018learning,ivanov2018anonymous,Verma2018Graph,xinyi2018capsule} generate graph representations by globally fusing node features. 
PATCHY-SAN~\cite{Niepert2016Learning} represents a graph as a sequence of nodes and generates local normalized neighborhood representations for each node. 
RNN autoencoder based methods~\cite{taheri2018learning,ivanov2018anonymous} capture graph properties by sampling node sequence, which is implicit and unaware of exact graph structures. 
Graph capsule networks~\cite{Verma2018Graph,xinyi2018capsule} capture node features in the form of capsules and use a routing mechanism to generate high-level features. 
However, these methods are incapable of exploiting the hierarchical structure information of graphs. 

Local substructures (i.e., motifs and subgraphs) are exploited to capture more complex structural characteristics. 
Motif-based methods~\cite{monti2018motifnet,yang2018node,lee2019graph,peng2020motif} are limited to enumerate exact small structures within graphs as local structure features. 
The motif extraction rules must be manually designed with prior knowledge. 
Other works exploit subgraphs by motif combination~\cite{yang2018node}, subgraph isomorphism counting~\cite{bouritsas2020improving}, and rule-based extraction~\cite{xuan2019subgraph} for graph-level tasks (e.g., subgraph classification~\cite{alsentzer2020subgraph}, graph evolution prediction~\cite{meng2018subgraph}, and graph classification~\cite{yang2018node,bouritsas2020improving,xuan2019subgraph}). 
NEST~\cite{yang2018node} explores subgraph-level patterns by various combinations of motifs. 
The most relevant work to ours is SGN~\cite{xuan2019subgraph}, which detects and selects appropriate subgraphs based on pre-defined rules, expanding the structural feature space effectively. 
Compared to our model, the subgraph detection and selection procedure is based on heuristics, and it is difficult for SGN to provide sufficient information when subgraphs become too large.

The aforementioned methods have many limitations in terms of discrimination, prior knowledge, and interpretability. 
In our framework, we address these problems by representing graphs as adaptively selected striking subgraphs. 

\begin{figure*}[t]
\centerline{\includegraphics[width=0.97\linewidth]{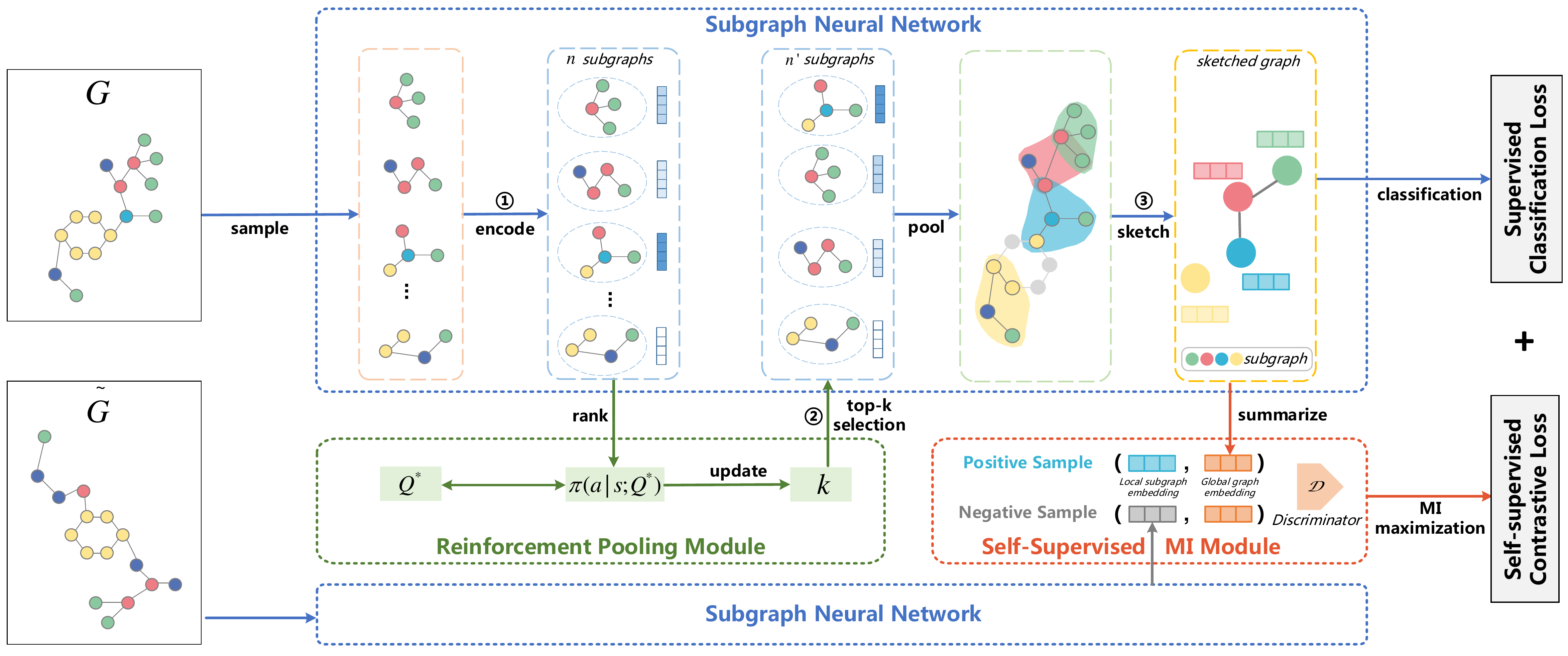}}
\caption{
An illustration of the \projtitle~architecture. 
Assuming the single graph setup (i.e., $G$ is provided as input and $\tilde{G}$ is an alternative graph providing negative samples), \projtitle~consists of the following steps: 
\textcircled{1} Subgraph sampling and encoding: for each graph, a fixed number of subgraphs is sampled and encoded by an intra-subgraph attention mechanism; 
\textcircled{2} Subgraph selection: striking subgraphs are selected by a reinforcement learning module and pooled into a sketched graph; 
\textcircled{3} Subgraph sketching: every supernode (i.e., subgraph) in the sketched graph is fed into an inter-subgraph attention layer; 
Subgraph representations are further enhanced by maximizing mutual information between local subgraph (in \textcolor{cyan}{cyan}) and global graph (in \textcolor{orange}{orange}) representations; 
the graph classification result is voted by classifying subgraphs.
}
\label{fig:framework}
\end{figure*}

\subsection{Graph Pooling}
Graph pooling is investigated to reduce the entire graph information into a coarsened graph, which broadly falls into two categories: cluster pooling and top-k selection pooling. 

\textit{Cluster pooling methods} (e.g., DiffPool~\cite{ying2018hierarchical}, EigenPooling~\cite{ma2019graph} and ASAP~\cite{ranjan2020asap}) group nodes into clusters and coarsen the graph based on the cluster assignment matrix. 
Feature and structure information is utilized implicitly during clustering, leading to a lack of interpretability. 
Furthermore, the number of clusters is always determined by a heuristic or performs as a hyper-parameter, and cluster operations always lead to high computational cost. 

\textit{Top-k selection pooling methods} compute the importance scores of nodes and select nodes with pooling ratio $k$ to remove redundant information. 
Top-k pooling methods are generally more memory efficient as they avoid generating dense cluster assignments. 
\cite{cangea2018towards} and \cite{gao2019graph} select nodes based on their scalar projection values on a trainable vector. 
SortPooling~\cite{zhang2018end} sorts nodes according to their structural roles within the graph using the WL kernel. 
SAGPool~\cite{lee2019self} uses binary classification to decide the preserving nodes. 
To our best knowledge, current top-k selection pooling methods are mostly based on heuristics, since they cannot parameterize the optimal pooling ratio~\cite{lee2019self}. 
The pooling ratio is always taken as a hyper-parameter and tuned during the experiment~\cite{lee2019self,gao2019graph}, which lacks a generalization ability. 
However, the pooling ratios are diverse in different types of graphs and should be chosen adaptively. 

Our framework adopts a reinforcement learning algorithm to optimize the pooling ratio, which can be trained with graph neural networks in an end-to-end manner. 

\subsection{Self-Supervised Learning on Graphs}
Self-supervised learning has shown superiority in boosting the performance of many downstream applications in computer vision~\cite{hjelm2019learning,he2020momentum} and natural language processing~\cite{devlin2018bert,yang2019xlnet}. 
Recent works~\cite{velickovic2019deep,peng2020graph,sun2020infograph,qiu2020gcc} harness self-supervised learning for GNNs and have shown competitive performance. 
DGI~\cite{velickovic2019deep} learns a node encoder that maximizes the mutual information between patch representations and corresponding high-level summaries of graphs. 
GMI~\cite{peng2020graph} generalizes the conventional computations of mutual information from vector space to the graph domain, where measuring the mutual information from the two aspects of node features and topological structure. 
InfoGraph~\cite{sun2020infograph} learns graph embeddings by maximizing the mutual information between graph embeddings and substructure embeddings of different scales (e.g., nodes, edges, triangles). 
Our approach differs in that we aim to obtain subgraph-level representations mindful of the global graph structural properties. 

\section{Our Approach}
This section proposes the framework of \projtitle~for graph classification, 
addressing the challenges of discrimination, prior knowledge, and interpretability simultaneously. 
The overview of \projtitle~is shown in Figure~\ref{fig:framework}. 
We first introduce the subgraph neural network, and then followed by the reinforcement pooling mechanism and the self-supervised mutual information mechanism.

We represent a graph as $G=(V,X,A)$, where $V=\{v_1,v_2,\cdots,v_N\}$ denotes the node set, $X\in\mathbb{R}^{N \times d}$ denotes the node features, and $A\in\mathbb{R}^{N \times N}$ denotes the adjacency matrix. 
$N$ is the number of nodes, and $d$ is the dimension of the node feature. 
Given a dataset $(\mathcal{G, Y})=\{(G_1, y_1), (G_2, y_2), \cdots (G_n, y_n)\}$, where $y_{i}\in\mathcal{Y}$ is the label of $G_{i}\in\mathcal{G}$, the task of graph classification is to learn a mapping function $f:\mathcal{G}\rightarrow\mathcal{Y}$ that maps graphs to the label sets. 

\subsection{Subgraph Neural Network}\label{sec:subgraph}
As the main component of SUGAR, the subgraph neural network reconstructs a sketched graph by extracting striking subgraphs as the original graph's representative part to reveal subgraph-level patterns. 
In this way, the subgraph neural network preserves the graph properties through a three-level hierarchy: node, intra-subgraph, and inter-subgraph. 
Briefly, there are three steps to build a subgraph neural network: 
(1) sample and encode subgraphs from the original graph; 
(2) select striking subgraphs by a reinforcement pooling module; 
(3) build a sketched graph and learn subgraph embeddings by an attention mechanism and a self-supervised mutual information mechanism. 

\textbf{Step-1: Subgraph sampling and encoding}. 
First, we sample $n$ subgraphs from the original graph. 
We sort all nodes in the graph by their degree in descending order and select the first $n$ nodes as the central nodes of subgraphs. 
For each central node, we extract a subgraph using the breadth-first search (BFS) algorithm. 
The number of nodes in each subgraph is limited to $s$. 
The limitation of $n$ and $s$ is to maximize the original graph structure's coverage with a fixed number of subgraphs. 
Then, we obtain a set of subgraphs $\{g_1,g_2,\cdots,g_n\}$. 

Second, we learn a GNN-based encoder, $\mathcal{E}:\mathbb{R}^{s\times d}\times \mathbb{R}^{s\times s}\rightarrow\mathbb{R}^{s\times d_1}$, to acquire node representations within subgraphs, where $d_1$ is the dimension of node representation. 
Then the node representations $\mathbf{H}(g_{i})\in\mathbb{R}^{s\times d_1}$ for nodes in subgraph $g_{i}$ can be obtained by the generalized equation:
\begin{equation}
\label{eq:encoder}
    \mathbf{H}(g_{i})=\mathcal{E}(g_{i})=\{\mathbf{h}_{j}|v_{j}\in V(g_{i})\}.
\end{equation}
We unify the formulation of $\mathcal{E}$ as a message passing framework:
\begin{equation}
\label{eq:GNN}
 \mathbf{h}^{(l+1)}_{i}={\rm U}^{(l+1)}(\mathbf{h}^{(l)}_{i},{\rm AGG}({\rm M}^{(l+1)}(\mathbf{h}^{(l)}_{i},\mathbf{h}^{(l)}_{j})|v_{j}\in N(v_{i}))),
\end{equation}
where ${\rm M}(\cdot)$ denotes the message generation function, ${\rm AGG}(\cdot)$ denotes the aggregation function, and ${\rm U}(\cdot)$ denotes the state updating function. 
Various formulas of GNNs can be substituted for Eq.~\eqref{eq:encoder}. 

Third, to encode node representations into a unified subgraph embedding space, we leverage an intra-subgraph attention mechanism to learn the node importance within a subgraph. 
The attention coefficient $c^{(i)}_{j}$ for $v_{j}$ is computed by a single forward layer, indicating the importance of $v_{j}$ to subgraph $g_{i}$:
\begin{equation}
\label{eq:selfattentioncoefficient}
    c^{(i)}_{j} = \sigma(\mathbf{a}^{\rm{T}}_{intra}\mathbf{W}_{intra}\mathbf{h}^{(i)}_{j}),
\end{equation}
where $\mathbf{W}_{intra}\in\mathbb{R}^{d_1\times d_1}$ is a weight matrix, and $\mathbf{a}_{intra}\in\mathbb{R}^{d_1}$ is a weight vector. 
$\mathbf{W}_{intra}$ and $\mathbf{a}_{intra}$ are shared among nodes of all subgraphs. 
Then, we normalize the attention coefficients across nodes within a subgraph via a softmax function. 
After this, we can compute the representations $\mathbf{z}_{i}$ of $g_i$ as follows: 
\begin{equation}
\label{eq:selfattention}
    \mathbf{z}_{i} = \sum_{v_{j}\in \mathbf{V}(g_{i})}{c^{(i)}_{j}\mathbf{h}^{(i)}_{j}}.
\end{equation}

\textbf{Step-2: Subgraph selection. }
To denoise randomly sampled subgraphs, we need to select subgraphs with prominent patterns, typically indicated by particular subgraph-level features and structures. 
We adopt \textit{top-k} sampling with an adaptive pooling ratio $k\in(0,1]$ to select a portion of subgraphs. 
Specifically, we employ a trainable vector $\mathbf{p}$ to project all subgraph features to 1D footprints $\{val_{i}|g_{i}\in G\}$. 
$val_i$ measures how much information of subgraph $g_i$ can be retained when projected onto the direction of $\mathbf{p}$. 
Then, we take the $\{val_{i}\}$ as the importance values of subgraphs and rank the subgraphs in descending order. 
After that, we select the top $n'=\left \lceil k\cdot n\right \rceil$ subgraphs and omit all other subgraphs at the current batch. 
During the training phase, $val_i$ of subgraph $g_i$ on $\mathbf{p}$ is computed as: 
\begin{equation}
\label{eq:topk}
    \begin{matrix}
    val_i =\frac{\mathbf{z_{i}\mathbf{p}}}{\left \|\mathbf{p}\right \|},
        & 
    idx = rank(\{val_{i}\},n'),
\end{matrix}
\end{equation}
where $rank(\{val_{i}\},n')$ is the operation of subgraph ranking, which returns the indices of the $n'$-largest values in $\{val_{i}\}$. 
$idx$ returned by $rank(\{val_{i}\},n')$ denotes the indices of selected subgraphs. 
$k$ is updated every epoch by a reinforcement learning mechanism introduced in Section~\ref{sec:RL}. 

\textbf{Step-3: Subgraph sketching. }
Since we learned the independent representations of subgraph-level features, we assemble the selected subgraphs to reconstruct a sketched graph to capture their inherent relations. 
First, as shown in Fig.~\ref{fig:framework}, we reduce the original graph into a sketched graph $G^{ske}=(V^{ske},E^{ske})$ by treating the selected subgraphs as supernodes. 
The connectivity between supernodes is determined by the number of common nodes in the corresponding subgraphs. 
Specifically, an edge $e(i,j)$ will be added to the sketched graph when the number of common nodes in $g_i$ and $g_j$ exceeds a predefined threshold $b_{com}$. 
\begin{equation}
\begin{matrix}
    V^{ske}=\{g_{i}\}, \forall~i \in idx;  
    &
    E^{ske}=\{e_{i,j}\}, \forall~\left |V(g_{i})\bigcap V(g_{j}) \right |>b_{com}.
\end{matrix}
\end{equation}

Second, an inter-subgraph attention mechanism is adopted to learn the mutual influence among subgraphs from their vectorized feature. 
More specifically, the attention coefficient $\alpha_{ij}$ of subgraph $g_i$ on $g_j$ can be calculated by the multi-head attention mechanism as in~\cite{velivckovic2017graph}. 
Then the subgraph embeddings can be calculated as: 
\begin{equation}
\label{eq:sketching}
    \mathbf{z}'_{i} = \frac{1}{M}\sum^{M}_{m=1}\sum_{e_{ij}\in E^{ske}}\alpha^{m}_{ij}\mathbf{W}^{m}_{inter}\mathbf{z}_{i},
\end{equation}
where $\alpha_{ij}$ is the attention coefficient, $\mathbf{W}_{inter}\in \mathbb{R}^{d_2\times d_1}$ is a weight matrix, and $M$ is the number of independent attention. 

Third, the subgraph embeddings will be further enhanced by a self-supervised mutual information mechanism introduced in Section~\ref{sec:MI}. 

After obtaining the subgraph embeddings, we convert them to label prediction through a softmax function. 
The probability distribution on class labels of different subgraphs can provide an insight into the impacts of subgraphs on the entire graph. 
Finally, the graph classification results are voted by subgraphs. 
Concretely, the classification results of all the subgraphs are ensembled by applying sum operation as the final probability distribution of the graph. 
The indexed class with the maximum probability is assumed to be the predicted graph label. 

\subsection{Reinforcement Pooling Module}\label{sec:RL}
To address the challenge of prior knowledge in \textit{top-k} sampling, we present a novel reinforcement learning (RL) algorithm to update the pooling ratio $k$ adaptively, even when inputting subgraphs of varying sizes and structures. 
Since the pooling ratio $k$ in \textit{top-k} sampling does not directly attend the graph classification, it cannot be updated by backpropagation. 
We use an RL algorithm to find optimal $k\in \left ( 0, 1\right ]$ rather than tuning it as a hyper-parameter. 
We model the updating process of $k$ as a finite horizon Markov decision process (MDP). 
Formally, the state, action, transition, reward and termination of the MDP are defined as follows:
\begin{itemize}[leftmargin=*]
    \item \textit{State.} 
    The state $s_e$ at epoch $e$ is represented by the indices of selected subgraphs $idx$ defined in Eq.~\eqref{eq:topk} with pooling ratio $k$: 
    \begin{equation}
        s_e = idx_e
    \end{equation}
    \item \textit{Action.} 
    RL agent updates $k$ by taking action $a_e$ based on reward. 
    We define the action $a$ as add or minus a fixed value $\Delta k\in \left [ 0,1\right ]$ from $k$. 
    \item \textit{Transition.} 
    After updating $k$, we use \textit{top-k} sampling defined in Eq.~\eqref{eq:topk} to select a new set of subgraphs in the next epoch.
    \item \textit{Reward.} 
    Due to the black-box nature of GNN, it is hard to sense its state and cumulative reward. 
    So we define a discrete reward function $reward(s_e,a_e)$ for each $a_e$ at $s_e$ directly based on the classification results: 
    \begin{equation}
    \label{eq:reward}
        reward(s_e,a_e)=\left\{\begin{matrix}
        +1,& if~acc_{e}> acc_{e-1},\\ 
        0,& if~acc_{e}=acc_{e-1},\\ 
        -1,& if~acc_{e}<acc_{e-1}.
        \end{matrix}\right.
    \end{equation}
    where $acc_{e}$ is the classification accuracy at epoch $e$.  
    Eq.~\eqref{eq:reward} indicates if the classification accuracy with $a_e$ is higher than the previous epoch, the reward for $a_e$ is positive, and vice versa. 
    \item \textit{Termination.} 
    If the change of $k$ among ten consecutive epochs is no more than $\Delta k$, the RL algorithm will stop, and $k$ will remain fixed during the next training process. 
    This means that RL finds the optimal threshold, which can retain the most striking subgraphs. 
    The terminal condition is formulated as:
    \begin{equation}
    \label{eq:terminal}
        Range(\{k_{e-10},\cdots,k_{e}\})\leq \Delta k. 
    \end{equation}
    
\end{itemize}
Since this is a discrete optimization problem with a finite horizon, we use \textit{Q-learning}~\cite{watkins1992q} to learn the MDP. 
\textit{Q-learning} is an off-policy reinforcement learning algorithm that seeks to find the best action to take given the current state. 
It fits the Bellman optimality equation as follows: 
\begin{equation}
\label{eq:Bellman}
    Q^*(s_e,a_e)=reward(s_e,a_e)+\gamma \mathop{\arg\max}\limits_{a'}Q^*(s_{e+1},a'),
\end{equation}
where $\gamma\in [0,1]$ is a discount factor of future reward. 
We adopt a $\varepsilon$-\textit{greedy} policy with an explore probability $\varepsilon$:
\begin{equation}
\label{eq:action}
    \pi(a_{e}|s_e;Q^*)=\left\{\begin{matrix}
    \rm{random~action},& \rm{w.p.}~\varepsilon\\ 
    \mathop{\arg\max}\limits_{a_{e}}Q^*(s_e,a),& \rm{otherwise}
    \end{matrix}\right.
\end{equation}
This means that the RL agent explores new states by selecting an action at random with probability $\varepsilon$ instead of selecting actions based on the max future reward. 

The RL agent and graph classification model can be trained jointly in an end-to-end manner, and the complete process of the RL algorithm is shown in Lines~\ref{code:RLbegin}-\ref{code:RLend} of Algorithm~\ref{alg:1}. 
We have tried other RL algorithms such as multi-armed bandit and DQN, but their performance is not as good as the Q-learning algorithm. 
The experiment results in Section~\ref{sec:expRL} verify the effectiveness of the reinforcement pooling module.

\subsection{Self-Supervised Mutual Information Module}\label{sec:MI}
Since our model relies on extracting striking subgraphs as the representative part of the original graph, we utilize the mutual information (MI) to measure the expressive ability of the obtained subgraph representations. 
To discriminate the subgraph representations among graphs, we present a novel method that maximizes the MI between local subgraph representations and the global graph representation. 
All of the derived subgraph representations are constrained to to be mindful of the global structural properties, rather than enforcing the overarching graph representation to contain all properties.

To obtain the global graph representation $\mathbf{r}$, we leverage a READOUT function to summarize the obtained subgraph-level embeddings into a fixed length vector: 
\begin{equation}
\label{eq:readout}
    \mathbf{r} = {\rm READOUT}(\{\mathbf{z}'_{i}\}^{n'}_{i=1}).
\end{equation}
The READOUT function can be any permutation-invariant function, such as averaging and graph-level pooling. 
Specifically, we apply a simple averaging strategy as the READOUT function here. 

We use the Jensen-Shannon (JS) MI estimator~\cite{nowozin2016f} on the local/global pairs to maximize the estimated MI over the given subgraph/graph embeddings. 
The JS MI estimator has an approximately monotonic relationship with the KL divergence (the traditional definition of mutual information), which is more stable and provides better results~\cite{hjelm2019learning}. 
Concretely, a discriminator $\mathcal{D}: \mathbb{R}^{d_2}\times \mathbb{R}^{d_2}\rightarrow \mathbb{R}$ is introduced, which takes a subgraph/graph embedding pair as input and determines whether they are from the same graph. 
We apply a bilinear score function as the discriminator: 
\begin{equation}
\label{eq:discriminator}
    \mathcal{D}(\mathbf{z}'_{i}, \mathbf{r})=\sigma(\mathbf{z}'^{T}_{i}\mathbf{W}_{\scriptscriptstyle MI}\mathbf{r}),
\end{equation}
where 
$\mathbf{W}_{\scriptscriptstyle MI}$ is a scoring matrix and $\sigma(\cdot)$ is the sigmoid function. 

The self-supervised MI mechanism is contrastive, as our MI estimator is based on classifying local/global pairs and negative-sampled counterparts. 
Specifically, the negative samples are provided by pairing subgraph representation $\tilde{\mathbf{z}}$ from an alternative graph $\tilde{G}$ with $\mathbf{r}$ from $G$. 
As a critical implementation detail of contrastive methods, the negative sampling strategy will govern the specific kinds of structural information to be captured. 
In our framework, we take another graph in the batch as the alternative graph $\tilde{G}$ to generate negative samples in a batch-wise fashion. 
To investigate the impact of the negative sampling strategy, we also devise another MI enhancing method named \projtitleMICor, which samples negative samples in a corrupted graph (i.e. $\widetilde{G}(V,\widetilde{X},A)=\mathcal{C}(G(V,X,A))$). 
Following the setting in \cite{velickovic2019deep}, the corruption function $\mathcal{C}(\cdot)$ preserves original vertexes $V$ and adjacency matrix $A$, whereas it corrupts features, $\widetilde{X}$, by row-wise shuffling of $X$. 
We further analyze these two negative sampling strategies in Section~\ref{sec:expMI}. 

The self-supervised MI objective can be defined as a standard binary cross-entropy (BCE) loss: 
\begin{equation}
\label{eq:loss_MI}
\begin{aligned}
    \mathcal{L}^{\scriptscriptstyle G}_{\scriptscriptstyle MI} = \frac{1}{n'+n_{\scriptscriptstyle neg}}
    &(\sum^{n'}_{g_{i}\in G}\mathbb{E}_{\scriptscriptstyle pos}\left [ \log(\mathcal{D}(\mathbf{z}'_{i}, \mathbf{r}))\right ]+\\
    &\sum^{n_{\scriptscriptstyle neg}}_{g_{j}\in \tilde{G}}\mathbb{E}_{\scriptscriptstyle neg}\left [ \log(1-\mathcal{D}(\mathbf{\tilde{z}}'_{j}, \mathbf{r}))\right ]), 
\end{aligned}
\end{equation}
where $n_{\scriptscriptstyle neg}$ denotes the number of negative samples. 
The BCE loss $\mathcal{L}^{\scriptscriptstyle G}_{\scriptscriptstyle MI}$ amounts to maximizing the mutual information between $z'_{i}$ and $\mathbf{r}$ based on the Jensen-Shannon divergence between the joint distribution (positive samples) and the product of marginals (negative samples)~\cite{nowozin2016f,velickovic2019deep}. 
The effectiveness of the self-supervised MI mechanism and several insights are discussed in Section~\ref{sec:expMI}.

\subsection{Proposed \projtitle}
\textbf{Optimization. } 
We combine the purely supervised classification loss $\mathcal{L}_{\scriptscriptstyle Classify}$ and the self-supervised MI loss $\mathcal{L}^{\scriptscriptstyle G}_{\scriptscriptstyle MI}$ in Eq.~\eqref{eq:loss_MI}, which acts as a regularization term. 
The graph classification loss function $\mathcal{L}_{\scriptscriptstyle Classify}$ is defined based on cross-entropy. 
The loss $\mathcal{L}$ of \projtitle~is defined as follows:
\begin{equation}
\label{eq:loss}
    \mathcal{L} = \mathcal{L}_{\scriptscriptstyle Classify} + \beta\sum_{G\in\mathcal{G}}\mathcal{L}^{\scriptscriptstyle G}_{\scriptscriptstyle MI}+\lambda\left \| \Theta \right \|^{2},
\end{equation}
where $\beta$ controls the contribution of the self-supervised MI enhancement, and $\lambda$ is a coefficient for L2 regularization on $\Theta$, which is a set of trainable parameters in this framework. 
In doing so, the model is trained to predict the entire graph properties while keeping rich discriminative intermediate subgraph representations aware of both local and global structural properties. 

\textbf{Algorithm description. }
Since graph data in the real world are most large in scale, we employ the mini-batch technique in the training process. 
Algorithm~\ref{alg:1} outlines the training process of \projtitle. 

\begin{algorithm}[t]
\caption{The overall process of \projtitle}
\label{alg:1}
\LinesNumbered
\KwIn{Graphs with labels $\{G=(V,X,A), y\}$; Number of subgraphs $n$; Subgraph size $s$; Initialized pooling ratio $k_0$; Number of epochs, batches: $E$, $B$;}
\KwOut{Graph label $y$}
\tcp{Subgraph sampling}
Sort all nodes within a graph by their degree in descending order;\\
Extract subgraphs for the first $n$ nodes;\\
\tcp{Train \projtitle}
\For{$e=1,2,\cdots,E$}{
    \For{$b=1,2,\cdots,B$}{
        $\mathbf{H}(g_{i}), \forall g_{i}\in \mathcal{G}_{b}\leftarrow$ Eq.~\eqref{eq:encoder};\tcp*[f]{Subgraph encoding}\\
        $\mathbf{z}\leftarrow$ Eq.~\eqref{eq:selfattention};\tcp*[f]{Intra-subgraph attention}\\
        $idx\leftarrow$ Eq.~\eqref{eq:topk};\tcp*[f]{Subgraph selection}\\
        $G^{ske}\leftarrow G$;\tcp*[f]{Subgraph sketching}\\
        $\mathbf{z}'_{i}\leftarrow$ Eq.~\eqref{eq:sketching};\tcp*[f]{Inter-subgraph attention}\\
        \tcp{Self-Supervised MI}
        Sample negative samples;\\
        $\mathbf{r}\leftarrow $ Eq.~\eqref{eq:readout};\\
        $\mathcal{L}^{\scriptscriptstyle G}_{\scriptscriptstyle MI}\leftarrow$ Eqs.~\eqref{eq:discriminator} and \eqref{eq:loss_MI};\\
        $\mathcal{L}\leftarrow$ Eq.~\eqref{eq:loss};
    }
    \tcp{RL process}
    \If{Eq.~\eqref{eq:terminal} is False\label{code:RLbegin}}{
        $reward(s_e,a_e)\leftarrow$ Eq.~\eqref{eq:reward};\\
        $a_e\leftarrow$ Eq.~\eqref{eq:action};\\
        $k\leftarrow a_e\cdot\Delta k$;\label{code:RLend}
    }
}
\end{algorithm}

\section{Experiments}
In this section, we describe the experiments conducted to demonstrate the efficacy of \projtitle~for graph classification. 
The experiments aim to answer the following five research questions: 
\begin{itemize}[leftmargin=*]
    \item \textbf{Q1.} 
    How does \projtitle~perform in graph classification? 
    \item \textbf{Q2.} 
    How do the exact subgraph encoder architecture and subgraph size influence the performance of \projtitle? 
    \item \textbf{Q3.} 
    How does the reinforcement pooling mechanism influence the performance of \projtitle? 
    \item \textbf{Q4.} 
    How does the self-supervised mutual information mechanism influence the performance of  \projtitle? 
    \item \textbf{Q5.} 
    Does \projtitle~select subgraphs with prominent patterns and provide insightful interpretations? 
\end{itemize}
\subsection{Experimental Setups}
\textbf{Datasets.} 
We use six bioinformatics datasets namely MUTAG~\cite{debnath1991structure}, PTC~\cite{Toivonen2003Statistical},  PROTEINS~\cite{borgwardt2005protein},  D\&D~\cite{Dobson2003Distinguishing},  NCI1~\cite{Wale2008Comparison}, and  NCI109~\cite{Wale2008Comparison}. 
The dataset statistics are summarized in Table~\ref{tab:dataset}. 
\begin{table}[ht!]
\caption{Statistics of Datasets. }
\label{tab:dataset}
\resizebox{\linewidth}{!}{%
\centering
\begin{tabular}{cccccc}
\hline
\textbf{Dataset} & \# Graphs & \# Classes & Max. Nodes & Avg. Nodes & Node Labels \\ \hline
MUTAG~\cite{debnath1991structure}            & 188       & 2          & 28         & 17.93      & 7           \\
PTC~\cite{Toivonen2003Statistical}              & 344       & 2          & 64         & 14.29      & 18          \\
PROTEINS~\cite{borgwardt2005protein}         & 1113      & 2          & 620        & 39.06      & 3           \\
D\&D~\cite{Dobson2003Distinguishing}             & 1178      & 2          & 5748       & 284.32     & 82          \\
NCI1~\cite{Wale2008Comparison}             & 4110      & 2          & 111        & 29.87      & 37          \\
NCI109~\cite{Wale2008Comparison}           & 4127      & 2          & 111        & 29.6       & 38          \\
\hline
\end{tabular}
}
\end{table}
\begin{table*}[ht]
\caption{Summary of experimental results: “average accuracy±standard deviation (rank)”. }
\centering
\label{tab:result}
\resizebox{0.96\linewidth}{!}{%
\begin{tabular}{ccccccccc}
\hline
\multirow{2}{*}{\textbf{Method}} & \multicolumn{6}{c}{\textbf{Dataset}}                                                                                                                         & \multirow{2}{*}{\textbf{Avg. Rank}} \\ \cline{2-7}
                                 & \textbf{MUTAG}         & \textbf{PTC}           & \textbf{PROTEINS}       & \textbf{D\&D}   & \textbf{NCI1}      & \textbf{NCI109}        &                                      \\ \hline
WL~\cite{shervashidze2011weisfeiler}                               & 82.05±0.36 (13)        & -                      & -                       & 79.78±0.36 (5)  & 82.19± 0.18 (6)    & 82.46±0.24 (3)            & 6.75                                \\
GK~\cite{Shervashidze2009Efficient}                               & 83.50±0.60 (12)        & 59.65±0.31 (9)         & -                       & 74.62±0.12 (13) & -                  & -                        & 11.33                               \\
DGK~\cite{Yanardag2015Deep}                              & 87.44±2.72 (9)         & 60.08±2.55 (8)         & 75.68±0.54 (12)         & -               & 80.31±0.46 (9)     & 80.32±0.33 (7)         & 9.00                                \\ \hline
PATCHY-SAN~\cite{Niepert2016Learning}                       & 92.63±4.21 (3)         & 62.29±5.68 (7)         & 75.89±2.76 (11)         & 77.12±2.41 (10) & 78.59±1.89 (10)    & -                      & 8.20                                \\
ECC~\cite{Simonovsky2017Dynamic}                              & 89.44 (6)              & -                      & -                       & 73.65 (14)      & 83.80 (2) & 81.87 (4)              & 6.50                                \\
GIN~\cite{xu2019powerful}                              & 89.40±5.60 (7)         & 64.60±7.00 (5)         & 76.20±2.80 (10)         & -               & 82.70±1.70 (5)     & -                      & 6.75                                \\
GCAPS-CNN~\cite{Verma2018Graph}                        & -                      & 66.01±5.91 (4)         & 76.40±4.17 (7)          & 77.62±4.99 (9)  & 82.72±2.38 (4)     & 81.12±1.28 (6)         & 6.00                                \\
CapsGNN~\cite{xinyi2018capsule}                          & 86.67±6.88 (10)        & -                      & 76.28±3.63 (8)          & 75.38±4.17 (12) & 78.35±1.55 (11)    & -                      & 10.25                                \\
AWE~\cite{ivanov2018anonymous}                              & 87.87±9.76 (8)         & -                      & -                       & 71.51±4.02 (15) & -                  & -                      & 11.50                               \\
S2S-N2N-PP~\cite{taheri2018learning}                       & 89.86±1.10 (5)         & 64.54±1.10 (6)         & 76.61±0.50 (4)          & -               & 83.72±0.40 (3)     & 83.64±0.30 (2)         & 4.00                                \\
NEST~\cite{yang2018node}                             & 91.85±1.57 (4)         & 67.42±1.83 (3)         & 76.54±0.26 (6)          & 78.11±0.36 (8)  & 81.59±0.46 (8)     & 81.72±0.41 (5)         & 5.67                                \\
MA-GCNN~\cite{peng2020motif}                          & 93.89±5.24 (2)         & 71.76±6.33 (2)         & 79.35±1.74 (2) & 81.48±1.03 (3)  & 81.77±2.36 (7)     & -                      & 3.20                                \\ \hline
SortPool~\cite{zhang2018end}                         & 85.83±1.66 (11)        & 58.59±2.47 (10)        & 75.54±0.94 (13)         & 79.37±0.94 (6)  & 74.44±0.47 (13)    & -                      & 10.60                               \\
DiffPool~\cite{ying2018hierarchical}                         & -                      & -                      & 76.25 (9)               & 80.64 (4)       & -                  & -                      & 6.50                                \\
gPool~\cite{gao2019graph}                            & -                      & -                      & 77.68 (3)               & 82.43 (2)       & -                  & -                      & 2.50                                \\
EigenPool~\cite{ma2019graph}                        & -                      & -                      & 76.60 (5)               & 78.60 (7)       & 77.00 (12)         & 74.90 (8)              & 8.00                                \\
SAGPool~\cite{lee2019self}                          & -                      & -                      & 71.86±0.97 (14)         & 76.45±0.97 (11) & 67.45±1.11 (14)    & 74.06±0.78 (9)         & 12.00                               \\ \hline
SUGAR (Ours)                     & \textbf{96.74±4.55(1)} & \textbf{77.53±2.82(1)} & \textbf{81.34±0.93(1)}            & \textbf{84.03±1.33(1)}    & \textbf{84.39±1.63(1)}      & \textbf{84.82±0.81(1)} & \textbf{1.00}                       \\ \hline
\end{tabular}
}
\end{table*}

\noindent\textbf{Baselines.}
We consider a number of baselines, including graph kernel based methods, graph neural network based methods, and graph pooling methods to demonstrate the effectiveness and robustness of \projtitle. 
\textit{Graph kernel based baselines} include Weisfeiler-Lehman Subtree Kernel (WL)~\cite{shervashidze2011weisfeiler}, Graphlet kernel (GK)~\cite{Shervashidze2009Efficient}, and 
Deep Graph Kernels (DGK)~\cite{Yanardag2015Deep}. 
\textit{Graph neural network based baselines} include PATCHY-SAN~\cite{Niepert2016Learning}, 
Dynamic Edge CNN (ECC)~\cite{Simonovsky2017Dynamic}, GIN~\cite{xu2019powerful}, 
Graph Capsule CNN (GCAPS-CNN)~\cite{Verma2018Graph}, CapsGNN~\cite{xinyi2018capsule}, 
Anonymous Walk Embeddings (AWE)~\cite{ivanov2018anonymous}, 
Sequence-to-sequence Neighbors-to-node Previous Predicted (S2S-N2N-PP)~\cite{taheri2018learning}, 
Network Structural Convolution (NEST)~\cite{yang2018node}, and MA-GCNN~\cite{peng2020motif}. 
\textit{Graph pooling baslines} include SortPool~\cite{zhang2018end},  DiffPool~\cite{ying2018hierarchical}, gPool~\cite{gao2019graph}, EigenPooling~\cite{ma2019graph}, and SAGPool~\cite{lee2019self}. 
\\
\noindent\textbf{Parameter settings.} 
The common parameters for training the models are set as $Momentum=0.9$, $Dropout=0.5$, 
and L2 norm regularization weight decay = 0.01. 
Node features are one-hot vectors of node categories. 
We adopt GCN~\cite{kipf2016semi} with 2 layers and 16 hidden units as our subgraph encoder. 
Subgraph embedding dimension $d'$ is set to 96. 
In the reinforcement pooling module, we set $\gamma=1$ in~\eqref{eq:Bellman} and $\varepsilon=0.9$ in~\eqref{eq:action}. 
For each dataset, the parameters $n, s, m, \Delta k$ are set based on the following principles: 
(1) Subgraph number $n$ and subgraph size $s$ are set based on the average size of all graphs; 
(2) $n_{\scriptscriptstyle neg}$ is set to the same value as $n'$; 
(3) $\Delta k$ is set to $\frac{1}{n}$. 

\subsection{Overall Evaluation (Q1)}
In this subsection, we evaluate \projtitle~for graph classification on the aforementioned six datasets. 
We performed 10-fold cross-validation on each of the datasets. 
The accuracies, standard deviations, and ranks are reported in Table~\ref{tab:result} where the best results are shown in bold. 
The reported results of the baseline methods come from the initial publications (“–” means not available).

As shown in Table~\ref{tab:result}, \projtitle~consistently outperforms all baselines on all datasets. 
In particular, \projtitle~achieves an average accuracy of 96.74\% on the MUTAG dataset, which is a 3.04\% improvement over the second-best ranked method MA-GCNN~\cite{peng2020motif}. 
Compared to node selection pooling baselines (e.g., gPool~\cite{gao2019graph}, SAGPool~\cite{lee2019self}), \projtitle~achieves more gains consistently, supporting the intuition behind our subgraph-level denoising approach. 
Compared to the recent hierarchical method NEST~\cite{yang2018node} and motif-based method MA-GCNN~\cite{peng2020motif}, our method achieves 14.99\% and 8.04\% improvements in terms of average accuracy on the PTC dataset, respectively. 
This may be because that both of NEST~\cite{yang2018node} and MA-GCNN~\cite{peng2020motif} are limited in their ability to enumerated simple motifs, while our method can capture more complex structural information by randomly sampling rather than by pre-defined rules. 

\textit{Overall, the proposed \projtitle~shows very promising results against recently developed methods. }

\subsection{Subgraph Encoder and Size Analysis (Q2)}
\label{sec:encoder}
\begin{figure}
    \centering
    \includegraphics[width=0.96\linewidth]{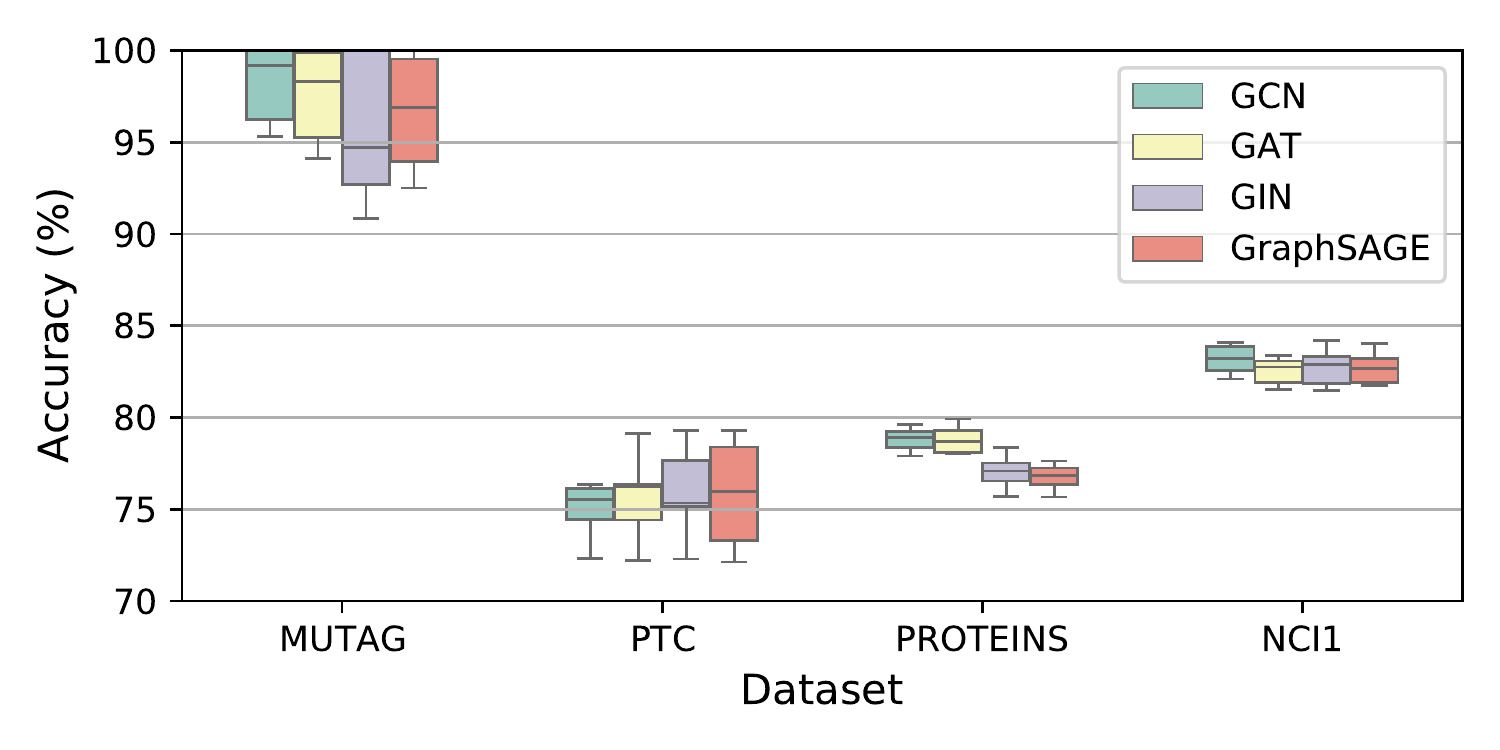}
    \caption{\projtitle~with different encoder architecture.}
    \label{fig:encoder}
\end{figure}

\begin{figure}
    \centering
    \includegraphics[width=0.96\linewidth]{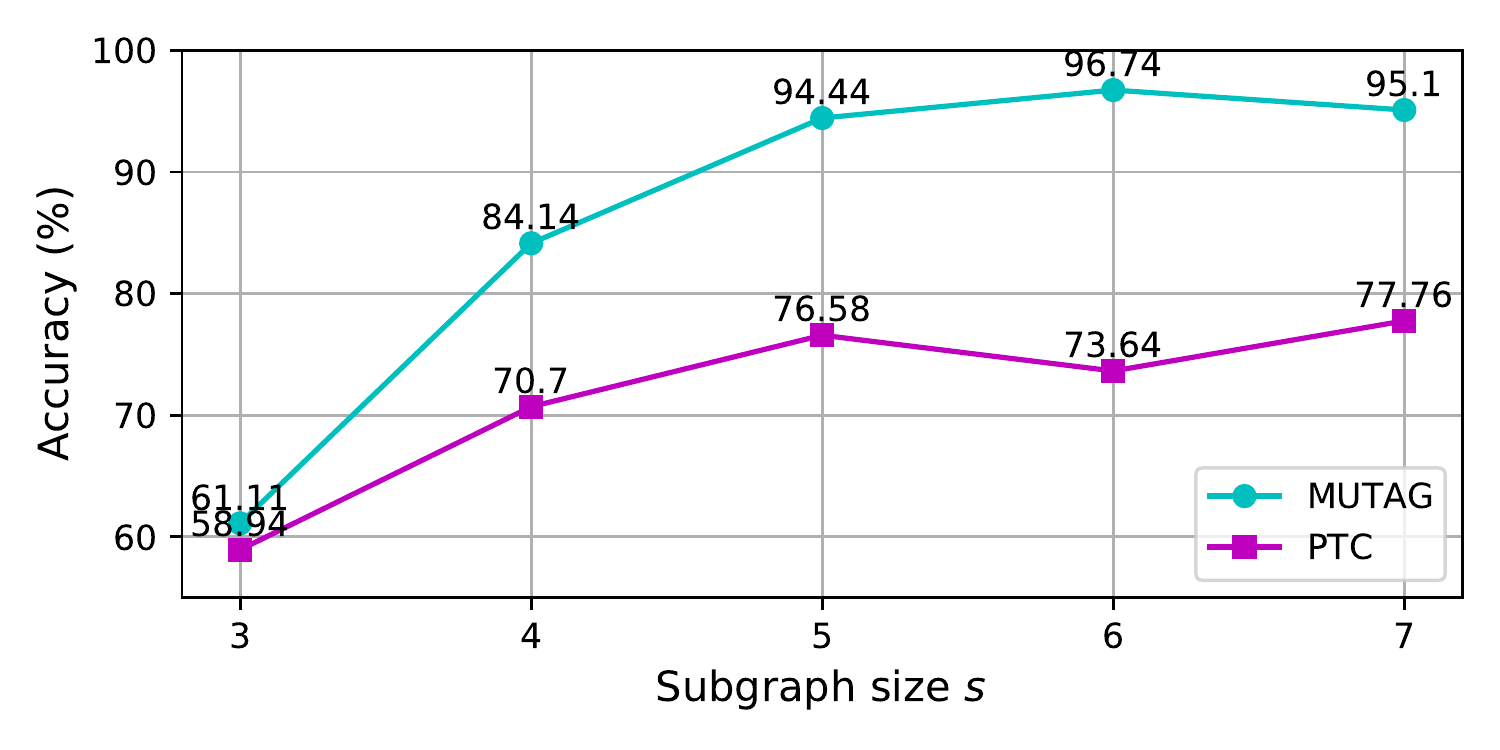}
    \caption{\projtitle~with different subgraph size $s$.}
    \label{fig:size}
\end{figure}

\begin{figure*}[htb]
\centering
\subfigure[Training process of \projtitleRL~and \projtitle~on PTC.]{
\begin{minipage}[t]{ 0.32\linewidth}
\centering
\includegraphics[width=\linewidth]{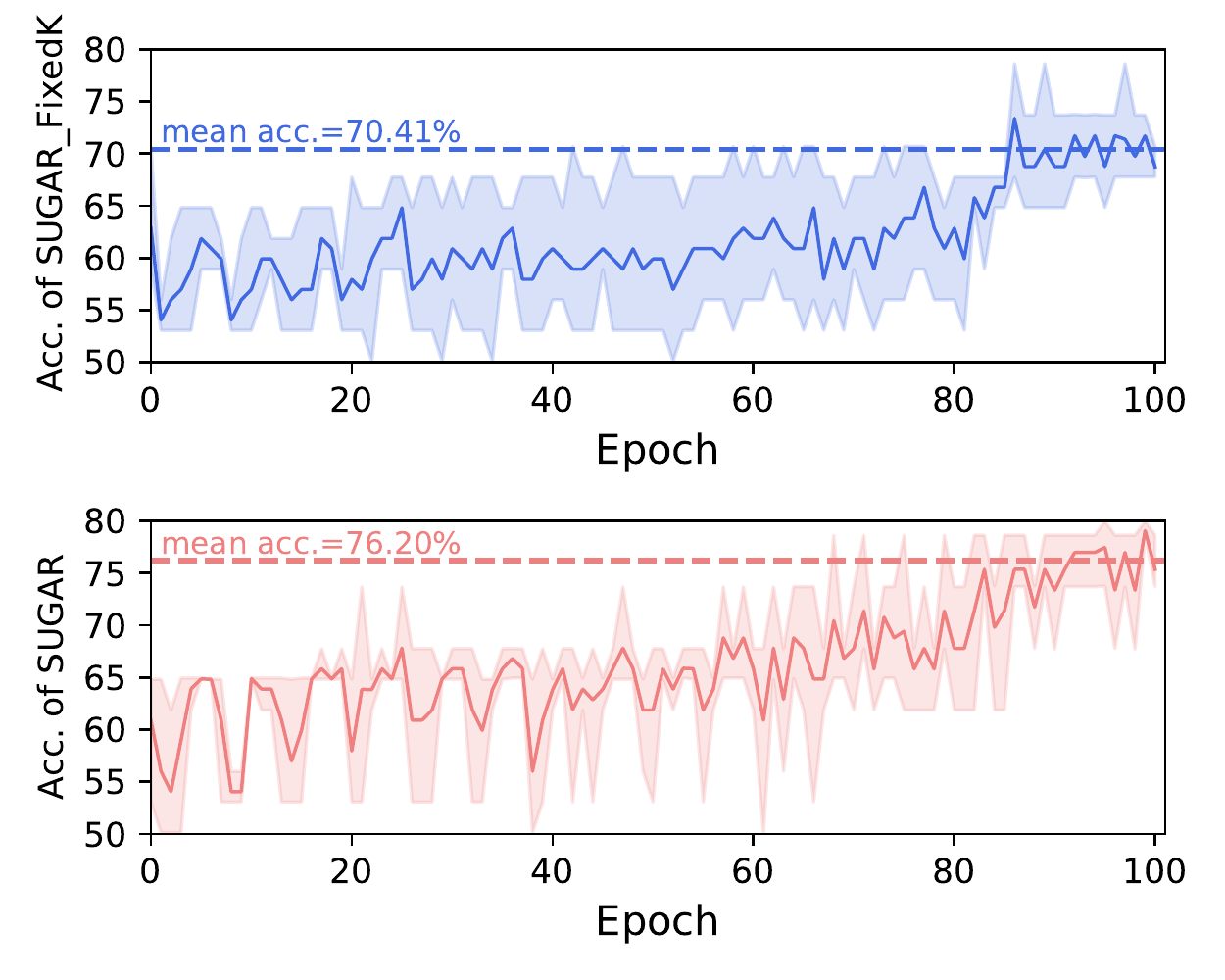}
\label{fig:nonRL}
\end{minipage}
}
\subfigure[Updating process of $k$ on PTC.]{
\begin{minipage}[t]{ 0.32\linewidth}
\centering
\includegraphics[width=\linewidth]{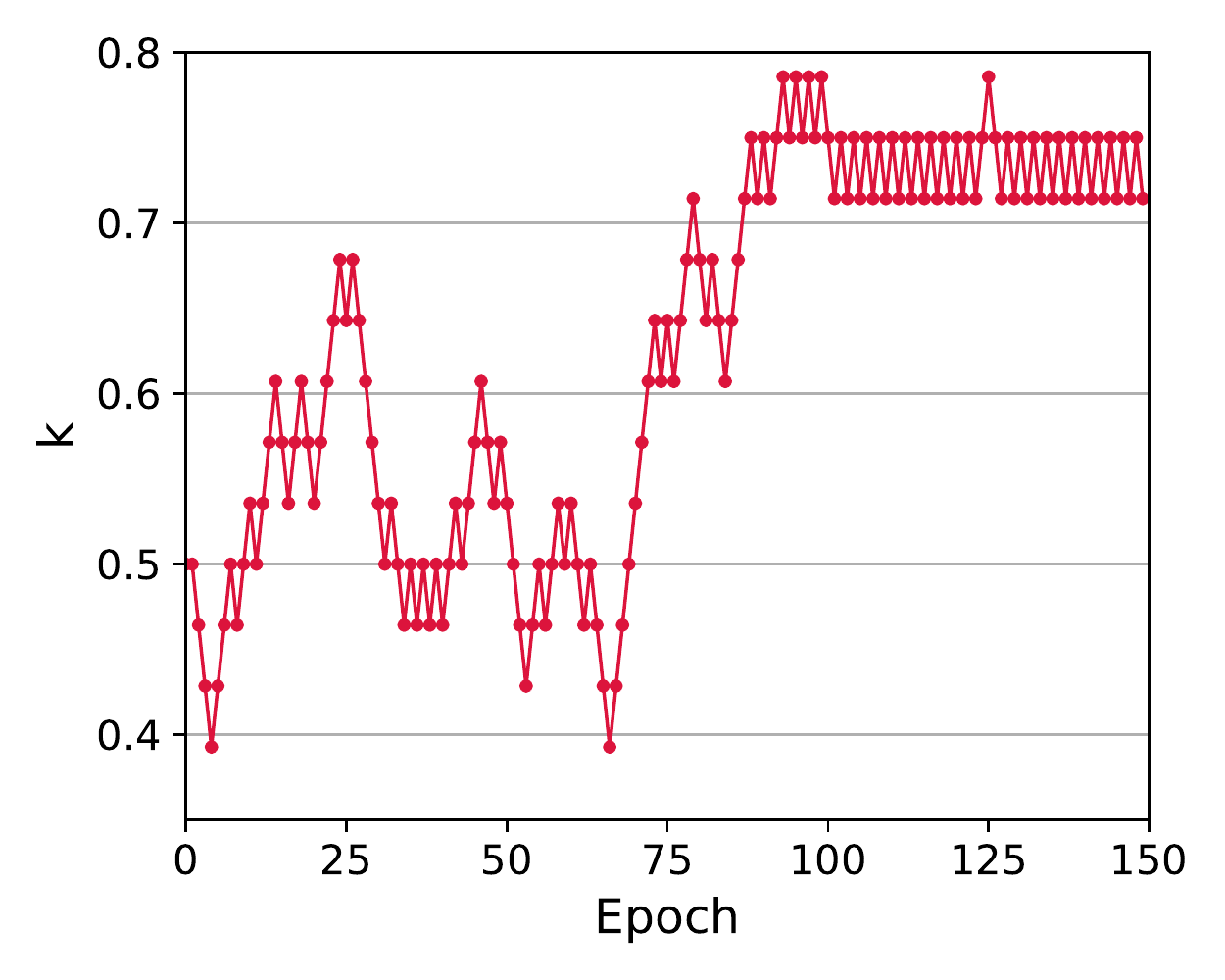}
\label{fig:k}
\end{minipage}
}
\subfigure[Learning curve of RL on PTC.]{
\begin{minipage}[t]{ 0.32\linewidth}
\centering
\includegraphics[width=\linewidth]{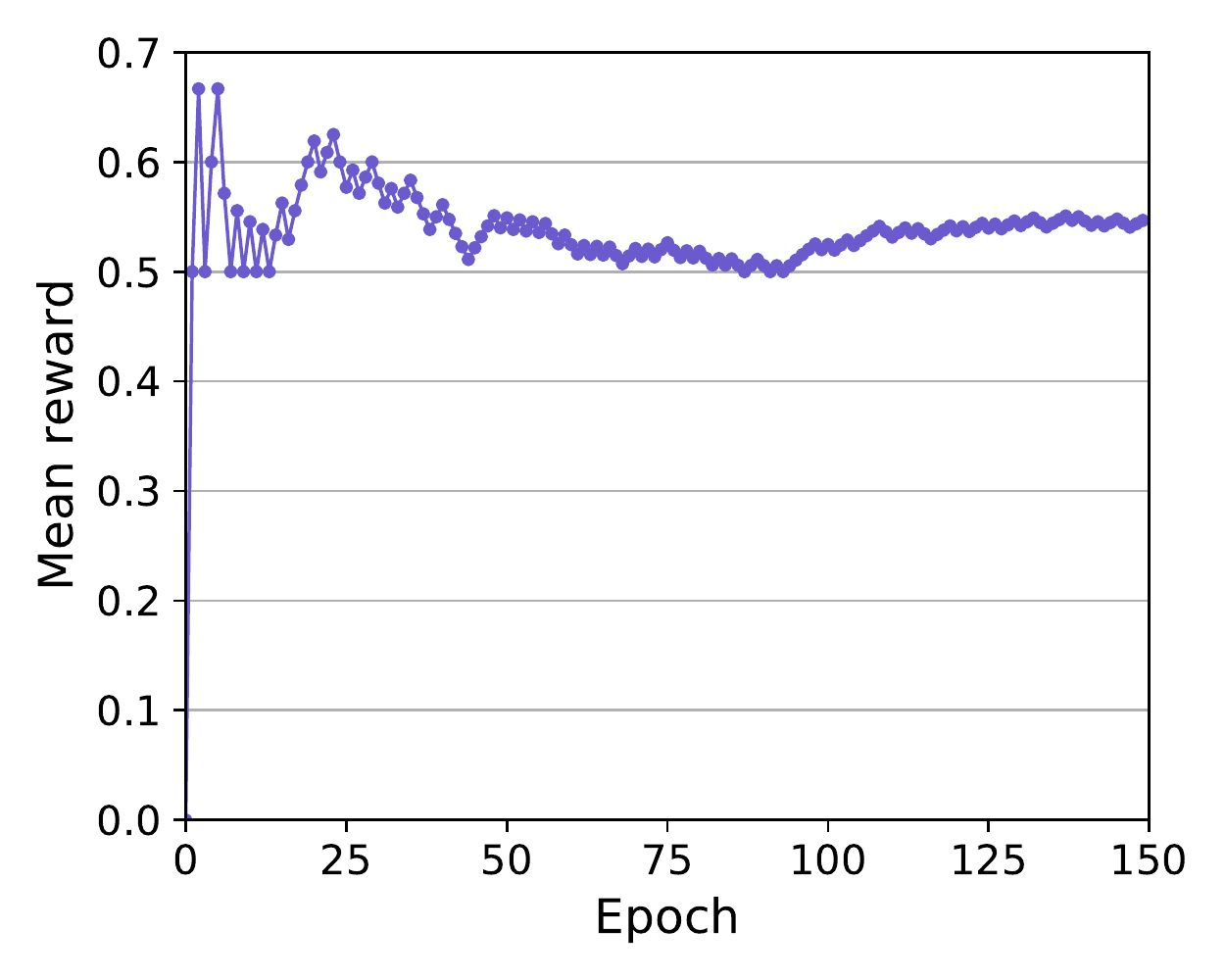}
\label{fig:reward}
\end{minipage}
}
\quad
\subfigure[\projtitle~with different negative sampling strategies.]{
\begin{minipage}[t]{ 0.32\linewidth}
\centering
\includegraphics[width=\linewidth]{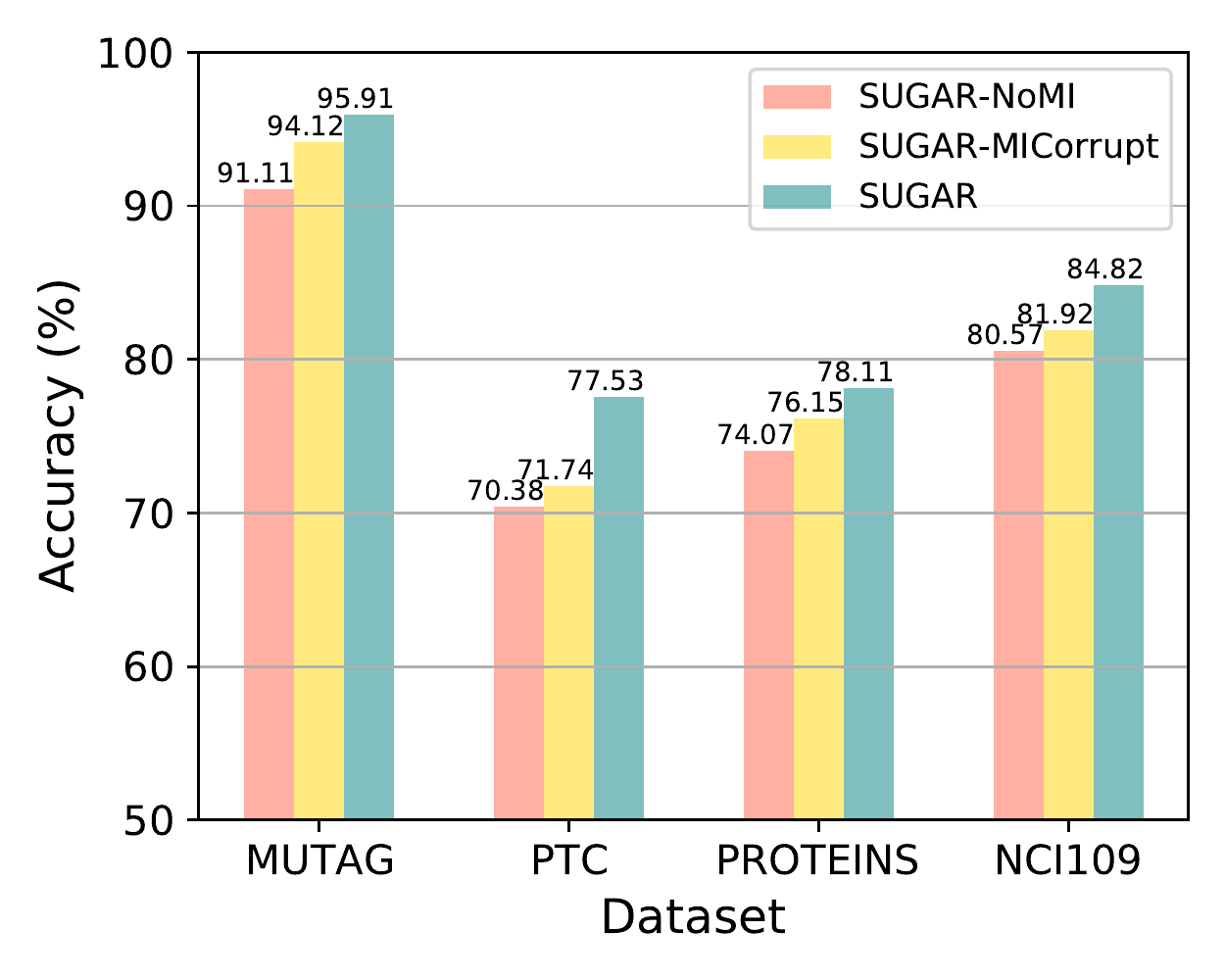}
\label{fig:MI}
\end{minipage}
}
\subfigure[Parameter sensitivity of negative sampling ratio.]{
\begin{minipage}[t]{ 0.32\linewidth}
\centering
\includegraphics[width=\linewidth]{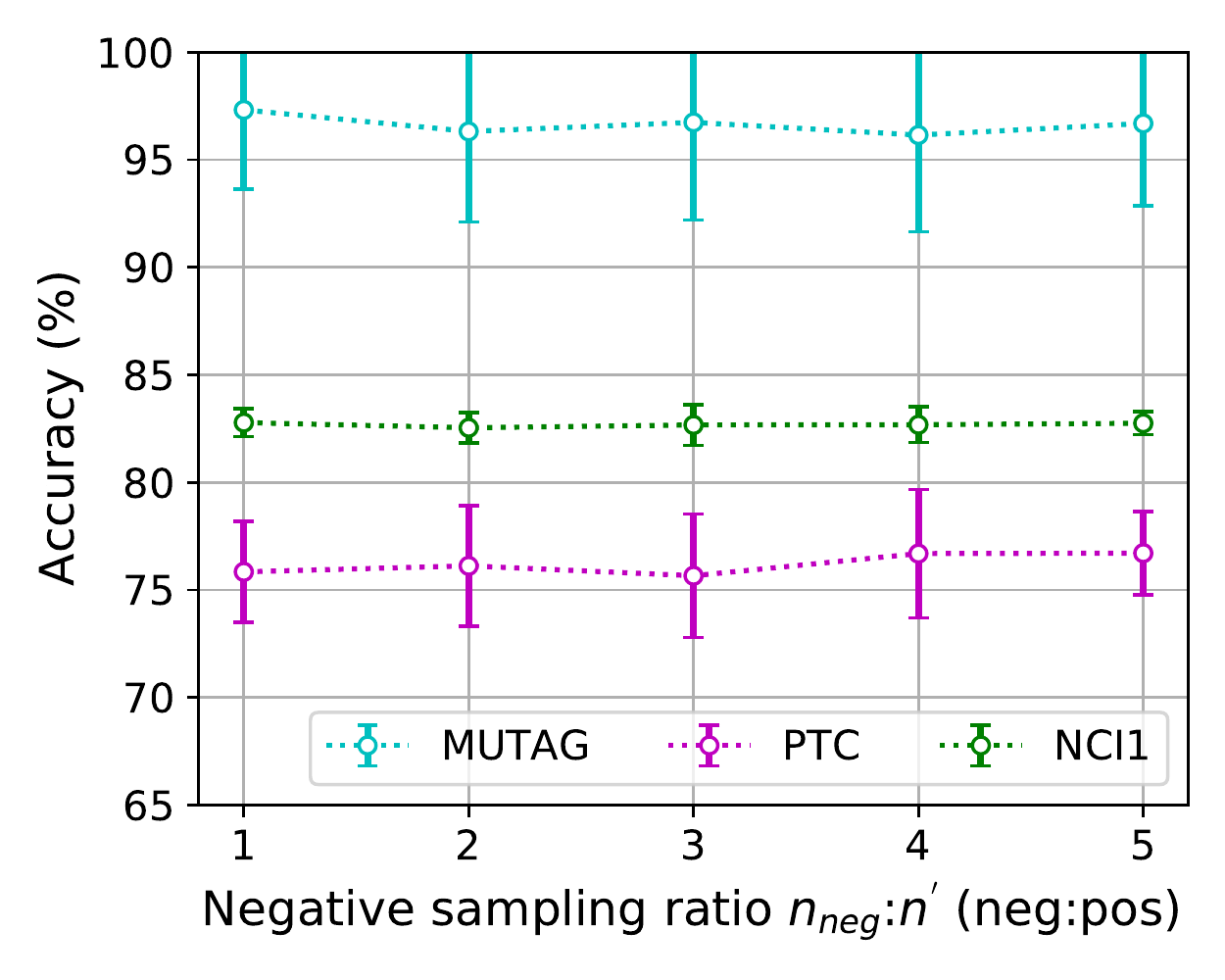}
\label{fig:neg_num}
\end{minipage}
}
\subfigure[Parameter sensitivity of MI loss coefficient $\beta$.]{
\begin{minipage}[t]{ 0.32\linewidth}
\centering
\includegraphics[width=\linewidth]{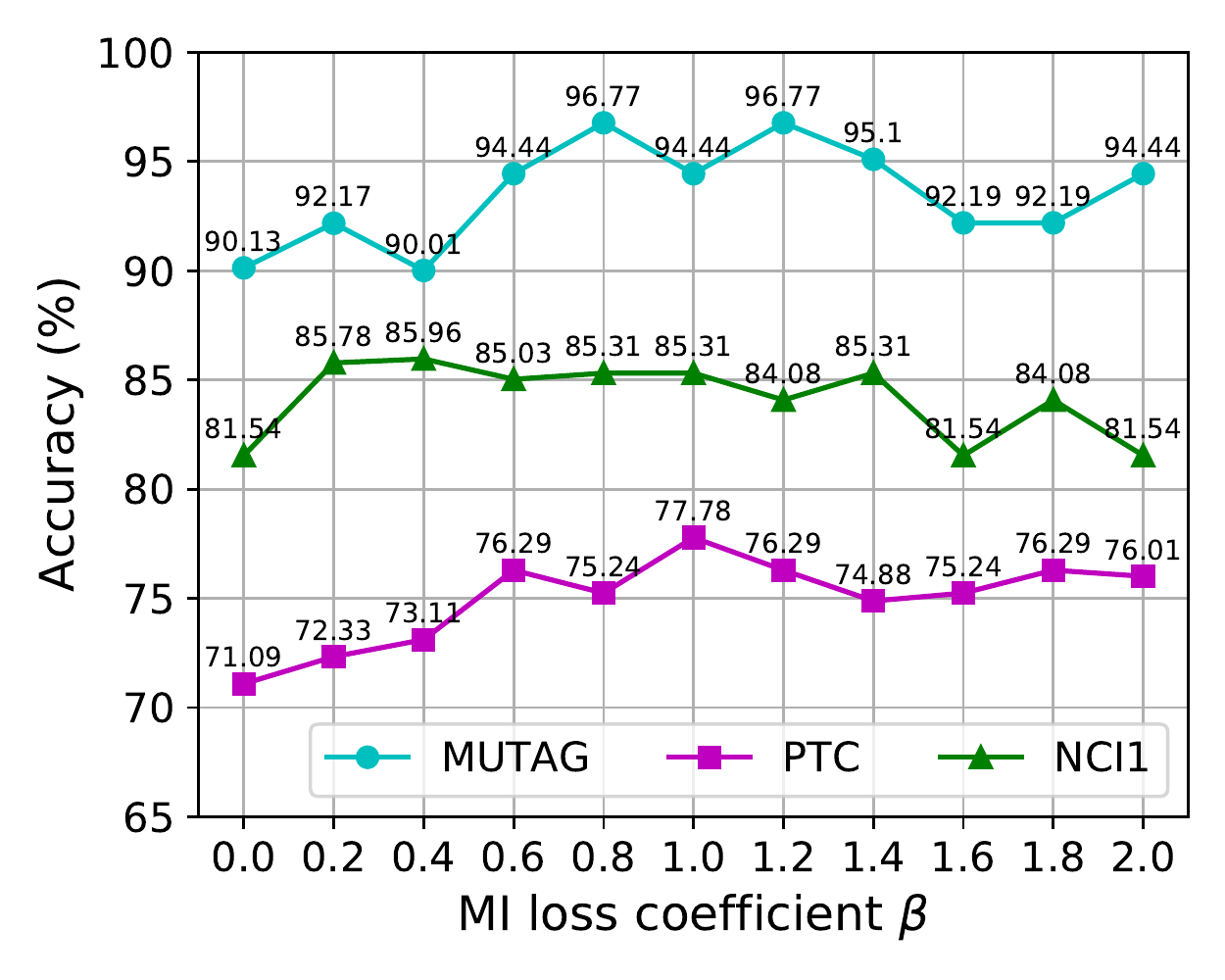}
\label{fig:MILoss}
\end{minipage}
}
\centering
\caption{The training process and testing performance of \projtitle. }
\label{fig:exp}
\end{figure*}
In this subsection, we analyze the impacts of subgraph encoder architecture and subgraph size. 

As discussed in Section~\ref{sec:subgraph}, any GNN can be used as the subgraph encoder. 
In addition to using GCN~\cite{kipf2016semi} in the default experiment setting, we also perform experiments with three popular GNN architectures: 
GAT~\cite{velivckovic2017graph}, GraphSAGE\cite{hamilton2017inductive} (with mean aggregation), and GIN~\cite{xu2019powerful}. 
The results are summarized in Figure~\ref{fig:encoder}. 
We can observe that the performance difference resulting from the different GNNs are marginal. 
This may be because all the aforementioned GNNs are expressive enough to capture the subgraph properties. 
\textit{This indicates the proposed \projtitle~is robust to the exact encoder architecture. }



Figure~\ref{fig:size} shows the performance of \projtitle~with different subgraph size $s$ from 3 to 7 on MUTAG and PTC. 
Although our model does not give satisfactory results with subgraphs of 3 or 4 nodes, it is found that subgraphs of a larger size obviously help to improve the performance. 
We can also observe that the subgraph size does not significantly improve the performance of \projtitle~when it is larger than 5. 
\textit{This indicates that \projtitle~can achieve competitive performance when sampled subgraphs can cover most of the basic functional building blocks. }

\subsection{RL Process Analysis (Q3)}
\label{sec:expRL}
To verify the effectiveness of the reinforcement pooling mechanism, we plot the training process of \projtitle~(lower) and the variant \projtitleRL~(upper) on PTC in Figure~\ref{fig:nonRL}. 
\projtitleRL~removes the reinforcement pooling mechanism and uses a fixed pooling ratio $k=1$ (i.e., without subgraph selection). 
The shadowed area is enclosed by the min and max value of five cross-validation training runs. 
The solid line in the middle is the mean value of each epoch, and the dashed line is the mean value of the last ten epochs. 
The mean accuracy of \projtitle~with an adaptive pooling ratio achieves a 5.79\% improvement over \projtitleRL, supporting the intuition behind our subgraph denoising approach. 

Since the RL algorithm and the GNN are trained jointly, the updating and convergence process is indeed important. 
In Figure~\ref{fig:k}, we visualize the updating process of $k$ in PTC with the initial value $k_0=0.5$. 
Since other modules in the overall framework update with the RL module simultaneously, the RL environment is not very steady at the beginning. 
As a result, $k$ also does not update steadily during the first 70 epochs. 
When the framework gradually converges, $k$ bumps for several rounds and meets the terminal condition defined in Eq.~\eqref{eq:terminal}. 
Figure~\ref{fig:reward} shows the learning curve in terms of mean reward. 
We can observe that the RL algorithm converges to the mean reward 0.545 with a stable learning curve. 

\textit{This suggests that the proposed \projtitle~framework can find the most striking subgraphs adaptively.}

\begin{figure*}[t]
\centerline{\includegraphics[width=0.98\linewidth]{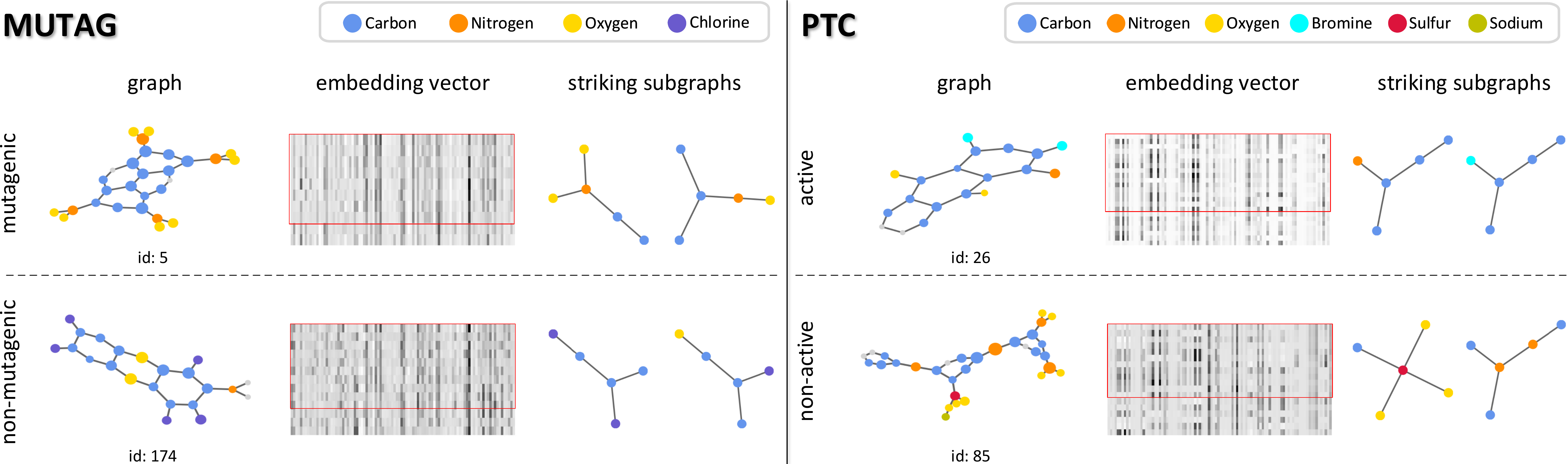}}
\caption{Result visualization of MUTAG (left) and PTC (right) dataset. }
\label{fig:visualization}
\end{figure*}
\subsection{Self-Supervised MI Analysis(Q4)}\label{sec:expMI}
In this subsection, we analyze the impact of the negative sampled strategy, the negative sampling ratio, and the sensitivity of the self-supervised MI loss coefficient. 

To evaluate the effectiveness of the self-supervised MI mechanism, we compare \projtitle~to its variant \projtitleMI, which removes the self-supervised MI mechanism. 
To further analyze the impact of the negative sampled strategy, we also compare \projtitle~to another variant \projtitleMICor, which constructs the alternative graph $\widetilde{G}$ by corruption as detailed in Section~\ref{sec:MI}. 
The results are shown in Fig.~\ref{fig:MI}. 
We can observe that models with the self-supervised MI mechanism (i.e., \projtitle~and \projtitleMICor) achieve better performance than \projtitleMI. 
In addition, \projtitle~(i.e., sampling from another graph instance) consistently outperforms than \projtitleMICor~(i.e., sampling from a corrupted graph). 
A possible reason is that $\widetilde{G}$ loses most of the important structure information during corruption and can only provide weak supervision. 

We also analyze the sensitivity of two hyper-parameters in the self-supervised MI module, namely negative sampling ratio $(n_{\scriptscriptstyle neg}:n')$ and the coefficient of self-supervised MI loss $\beta$. 
The common belief is that contrastive methods require a large number of negative samples to be competitive. 
Figure~\ref{fig:neg_num} shows the performance of \projtitle~under different negative sampling ratios. 
The larger negative sampling ratio does not seem to contribute significantly to boosting the performance of \projtitle. 
This may be because we draw negative samples for every subgraph within the graph. 
Though the negative sampling ratio is small, it has already provided sufficient self-supervision. 
As shown in Figure~\ref{fig:MILoss}, when the self-supervised MI loss has more than 0.6 weights compared to graph classification loss, \projtitle~achieves better performance. 
This illustrates that our framework quite benefits from self-supervised training. 

\textit{This indicates that MI enhancement gives informative self-supervision to \projtitle~and negative sampling strategy designs should be considered carefully. 
}

\subsection{Visualization (Q5)}\label{sec:visual}
In this subsection, we study the power of \projtitle~to discover subgraphs with prominent patterns and provide insightful interpretations into the formation of different graphs. 
Figure~\ref{fig:visualization} illustrates some of the results we obtained on the MUTAG (left) and PTC (right) dataset, where id denotes the graph index in the corresponding dataset. 
Each row shows our explanations for a specific class in each dataset. 
Column 1 shows a graph instance after subgraph selection, where the color indicates the atom type of the node (nodes in grey are omitted during subgraph selection) and size indicates the importance of the node in discriminating the two classes. 
Column 2 shows the $n\times 32$ neuron outputs in descending order of their projection value in the reinforcement pooling module.
The first $n'$ rows in the neuron output matrix with the largest projection value are selected as striking subgraphs, roughly indicating their activeness. 
Column 3 shows striking subgraphs as functional blocks found by \projtitle. 

MUTAG consists of 188 molecule graphs labeled according to whether the compound has a mutagenic effect on a bacterium. 
We observe that the main determinants in the mutagenic class is the nitro group $NO_2$ connected to a set of carbons. 
This is the same as the results in~\cite{debnath1991structure} which show that the electron-attracting elements conjugated with nitro groups are critical to identifying mutagenic molecules. 
For the non-mutagenic class, our model takes chlorine connected to carbons as a striking subgraph, which is frequently seen in non-mutagenic molecules. 

PTC consists of 344 organic molecules labeled according to whether the compound has carcinogenicity on male rats. 
The main determinants found by our model are the co-occurrence of carbon rings, nitrogen, sulfur, and oxygen. 
For instance, one of the striking subgraphs in active compounds is a nitrogen connected to a set of aromatic carbon bonds. 
This substructure is frequently seen in aromatic amines, nitroaromatics, and azo compounds, which are well-known classes of  carcinogens~\cite{Toivonen2003Statistical}. 
In addition, our model takes bromine connected to some carbons as a striking subgraph, which is in general agreement with accepted toxicological knowledge. 
For the non-active class, a striking subgraph found by our model is some oxygen with sulphur bond, which is the same as the knowledge of Leuven2~\cite{debnath1991structure}. 

\textit{This suggests that the proposed \projtitle~can find striking subgraphs with discriminative patterns and has great promise to provide sufficient interpretability. 
}
\section{Conclusion and Future Works}
In this paper, we proposed \projtitle, a novel end-to-end graph classification framework by subgraph selection and representation, addressing the challenges of discrimination, prior knowledge, and interpretability. 
\projtitle~preserves both local and global properties hierarchically by reconstructing a sketched graph, selects striking subgraphs adaptively by a reinforcement pooling mechanism, and discriminates subgraph representations by a self-supervised mutual information maximization mechanism. 
Extensive experiments on graph classification show the effectiveness of our approach. 
The selected subgraphs and learned weights can provide good interpretability and in-depth insight into structural bioinformatics analysis. 
Future work includes relating the subgraph sampling strategy to the learned implicit type rules, adopting more advanced reinforcement learning algorithms, and investigating the multi-label problem of subgraphs. 
Applying \projtitle~to other complex datasets and applications such as subgraph classification is another avenue of future research. 

\begin{acks}
The corresponding author is Jianxin Li. 
The authors of this paper were supported by the NSFC through grants (U20B2053 and 61872022), ARC DECRA Project (No.DE200100964), State Key Laboratory of Software Development Environment (SKLSDE-2020ZX-12), NSF ONR N00014-18-1-2009, and NSF under grants (III-1763325, III-1909323, and SaTC-1930941). This work was also sponsored by CAAI-Huawei MindSpore Open Fund, and NSF of Guangdong Province through grant 2017A030313339. Thanks for computing infrastructure provided by Huawei MindSpore platform. 
\end{acks}

\bibliographystyle{ACM-Reference-Format}
\bibliography{ref}

\end{document}